# Predicting the Probability of Collision of a Satellite with Space Debris: A Bayesian Machine Learning Approach

João Simões Catulo[a]*, Cláudia Soares[b], Marta Guimarães[c]

[a] *Instituto Superior Técnico, Portugal*, joao.catulo@tecnico.ulisboa.pt
[b] *FCT-UNL, Portugal*, claudia.soares@fct.unl.pt
[c] *Neuraspace, Portugal*, marta.guimaraes@neuraspace.com
\* Corresponding Author

**Abstract**

Space is becoming more crowded in Low Earth Orbit due to increased space activity. Such a dense space environment increases the risk of collisions between space objects endangering the whole space population. Therefore, the need to consider collision avoidance as part of routine operations is evident to satellite operators. Current procedures rely on the analysis of multiple collision warnings by human analysts. However, with the continuous growth of the space population, this manual approach may become unfeasible, highlighting the importance of automation in risk assessment. In 2019, ESA launched a competition to study the feasibility of applying machine learning in collision risk estimation and released a dataset that contained sequences of Conjunction Data Messages (CDMs) in support of real close encounters. The competition results showed that the naive forecast and its variants are strong predictors for this problem, which suggests that the CDMs may follow the Markov property. The proposed work investigates this theory by benchmarking Hidden Markov Models (HMM) in predicting the risk of collision between two resident space objects by using one feature of the entire dataset: the sequence of the probability in the CDMs. In addition, Bayesian statistics are used to infer a joint distribution for the parameters of the models, which allows the development of robust and reliable probabilistic predictive models that can incorporate physical or prior knowledge about the problem within a rigorous theoretical framework and provides prediction uncertainties that nicely reflect the accuracy of the predicted risk. This work shows that the implemented HMM outperforms the naive solution in some metrics, which further adds to the idea that the collision warnings may be Markovian and suggests that this is a powerful method to be further explored.

## 1. Introduction

Since the beginning of the space age, with the launch of Sputnik-1 in 1957, the amount of resident space objects in Earth orbit has been steadily increasing, as shown in Figure *1*, which presents the evolution of the number of objects in space from 1957 until today.

The space environment is becoming progressively crowded and space traffic is undergoing notable changes fuelled by the development of commercial and private space activities and the deployment of large constellations, especially in the Low Earth Orbit (LEO) region, which further adds to the growth of space population.

This continuous growth of the number of objects in space can pose a great danger to all operational satellites since collisions between resident space objects create large amounts of fragments that are further released into orbit. These fragmentation events create numerous debris that are spread into different directions at different velocities and, over time, lead to a gradual pollution of a vast volume of space [2], eventually contaminating entire orbital altitudes. If measures are not taken, collisions between space objects can reach a cascading point, in which collisions may cause a sequence of new impacts, due to the high density of objects in orbit, posing a real threat to space missions and endangering the whole space population. This effect is known as the *Kessler Syndrome* [3]. For these reasons, the need to consider collision avoidance as part of routine operations is evident. Moreover, the collision probability estimation is seen as an essential task to protect active spacecraft from collision with other space objects, since it allows the operators to take informed decisions regarding the need of potential avoidance manoeuvres.

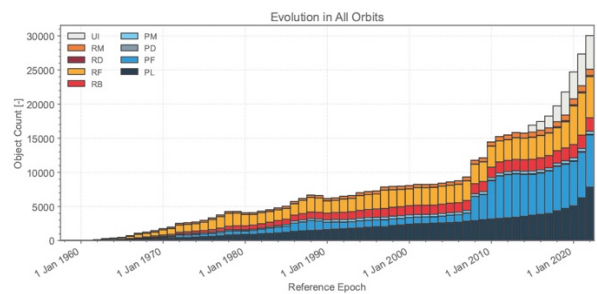

Figure 1. Count of objects in space from 1957 to 2022 [1].

When an event between two space objects meets the conditions for a close conjunction, in which the monitored space object is referred to as target and the other object as chaser, collision warnings in the form of conjunction data messages (CDMs) are created and sent to the operators of the satellites. These messages contain propagated information about the event to the time of closest approach (TCA). However, orbit determination and propagation cannot be modelled with desired precision and have associated uncertainties making it impossible to know for sure whether a collision will occur or not. Hence, during the time span of the conjunction event, both objects that generated the issue of warning messages are routinely tracked, leading to the creation of more CDMs that contain refined and more precise information about the conjunction. Typically, a LEO satellite receives hundreds of CDMs per week that, currently, require the analysis of human experts/analysts, generating high operational costs [4]. With the continuous growth of the space population, this approach may be an unfeasible task in the future, highlighting the importance of automation in risk assessment and estimation.

In 2019, the European Space Agency (ESA) launched the *Collision Avoidance Challenge* (CAC) [5] to study the feasibility of applying machine learning (ML) methods in collision risk estimation and released a dataset that contained sequences of CDMs received in support of real close encounters. The competition aimed to develop ML models capable of predicting the criticality of conjunction events by analysing the time series of CDMs received up to 2 days before the predicted TCA, which is considered the cut-off time. The collision probability within the CDMs is computed through the Alfriend-Akella algorithm [6] and the final risk of each event is considered to be the risk contained in the last released CDM, which is the best knowledge about the outcome of the close approach. Figure *2* illustrates the concept of ML in collision avoidance.

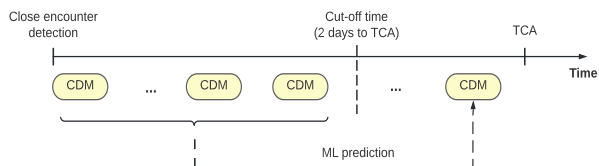

Figure 2. Concept of the ML approach in collision avoidance.

The competition showed that the naive/baseline approach (using the risk contained in the last CDM received until the cut-off time as the risk prediction) is a strong predictor for this problem, with only 12 teams out of 97 managing to beat the benchmark solution [7]. The team that presented the top solution used a step-by-step statistical approach to optimize the constitution of the test set and the competition metric. Manhattan LSTMs [8] and Gradient boosting trees showed good performance during the CAC.

After the competition, relevant work regarding the use of ML in collision avoidance has been conducted. Metz [9] implemented various models to predict the final chaser position uncertainties for each event and used those predictions to compute the risk using Akella's and Alfriend's algorithm [6]. Acciarini et al. [10] built a physics-based generative model using probabilistic programming to simulate the generation of CDMs, based on real data. Pinto et al. [11] used Bayesian deep learning with recurrent neural network architectures to also study the possibility of generating CDMs. Abay et al. [12] benchmarked the results for the state-of-the-art ML models that showed good results against the naive approach since the beginning of the competition.

*1.1 Objectives*

As mentioned, the naive forecast, as well as its variants, are very strong predictors for collision risk assessment, indicating that the time series of CDMs may follow the Markov property [7], i.e., the information contained in the current CDM only depends on the values of the previous CDM. In this work, this property will be investigated by implementing and benchmarking the use of hidden Markov model (HMM) in the risk prediction problem, using Bayesian statistics. For that, the dataset that ESA released for the CAC challenge is used.

*1.2 Why Hidden Markov Models?*

A HMM is a probabilistic model used to handle data which can be represented as a sequence of observations over time. It is a type of directed graphical model and a tool for representing probability distributions over sequences of observations that are produced by an underlying stochastic process, whose states cannot be directly observed, i.e., are hidden. This hidden process that generates the observations is a first-order finite state Markov chain and, hence, respects the Markov property that states that "*the probability distribution of future states of the process conditioned on both the past and present states depends only on the present state*" [13].

In the context of this work, in each event there is a physical process happening in space, in which the two objects in risk of colliding approach each other. This process cannot be observed and the CDMs can be interpreted as measurements that result from the physical approach. Thus, the hidden stochastic process of the HMM can be interpreted as the physical approach between the two objects, happening in space, and the observations can be seen as the CDMs.

2. **Background**

In this Section, the necessary theoretical concepts about Bayesian modelling and HMMs are provided. This section has been kept as brief as possible while giving all the necessary concepts. For more details, the reader is encouraged to read the references provided throughout this Section.

*2.1 Bayesian Modelling*

In probabilistic models, the set of parameters $\theta$ of a probabilistic model is typically obtained by finding the parameters that result in the best match between the model and the observed data **X**, using e.g., the maximum likelihood estimation. In this work, rather than estimating a single set of parameters, an entire joint distribution for $\theta$ is inferred. This is possible by adopting a Bayesian approach, in which the unknown parameters are treated as random variables and probability theory is used to update its values conditioned on the observed data [14]. The Bayesian interpretation considers that the associated randomness of $\theta$ encapsulates the prior belief one holds about the problem and that the belief is updated by some observed data **X**.

Bayesian modelling is based on Bayes' theorem, which states that

$$p(\theta|\mathbf{X}) = \frac{p(\mathbf{X}|\theta)\, p(\theta)}{p(\mathbf{X})}, \quad (1)$$

in which $p(\theta)$ denotes the prior distribution, $p(\mathbf{X}|\theta)$ the likelihood, $p(\mathbf{X})$ the evidence and $p(\theta|\mathbf{X})$ the posterior distribution. Once the posterior is defined, it can be used to obtain predictions of the model for new input data. However, computing the distribution $p(\theta|\mathbf{X})$ analytically is usually an unfeasible problem since it depends on the computation of the normalizing constant $p(\mathbf{X})$:

$$p(\mathbf{X}) = \int_\theta p(\mathbf{X}|\theta)\, p(\theta)\, d\theta, \quad (2)$$

where it is necessary to integrate over all the possible values of $\theta$. To address this issue, Markov chain Monte Carlo (MCMC) methods are used. These methods approximate the posterior distribution using samples, by evaluating the likelihood and prior distributions at different parameter values. In this work, Bayesian statistical models are implemented using a probabilistic programming framework called *PyMC* [15] and, to sample from the posterior, the No-U-Turn Sampler (NUTS) [16] is used.

*2.2 Hidden Markov Models*

As previously mentioned, a HMM is a probabilistic model used to represent probability distributions over sequences of observations that are produced by an underlying stochastic process that follows the Markov property and that cannot be directly observed. Throughout this work, the number of possible states that each latent variable can take will be denoted as *K*, the sequences of hidden states as $\mathbf{Z} = \{\mathbf{z}_1, \mathbf{z}_2, ..., \mathbf{z}_N\}$ and the sequences of observations as $\mathbf{X} = \{\mathbf{x}_1, \mathbf{x}_2, ..., \mathbf{x}_N\}$, in which each hidden state $\mathbf{z}_n$ generates the corresponding observation $\mathbf{x}_n$ (which may be of different type or dimension [17]) and N represents the number of observations.

A HMM with N observations is depicted as a graphical model in Figure *3*.

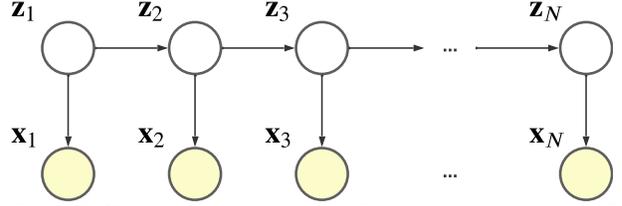

Figure 3. Graphical representation of a HMM for a sequence of N observations.

In HMMs, the latent variables **Z** follow a Markov chain with transition matrix $\mathbf{A} \in \mathbb{R}^{K \times K}: \mathbf{A} \geq 0, \mathbf{A}\mathbb{1} = \mathbb{1}$ (where $\mathbb{1}$ denotes a *K*-dimensional vector with elements equal to 1) and initial distribution $\boldsymbol{\pi} \in \mathbb{R}^K: \boldsymbol{\pi} \geq 0, \boldsymbol{\pi}^T\mathbb{1} = 1$ that represent the probability of transitioning from one hidden state to another – $p(\mathbf{z}_n|\mathbf{z}_{n-1})$ – and the hidden state initialization probability – $p(\mathbf{z}_1|\boldsymbol{\pi})$ – respectively. The *i*-th row of **A**, which will be denoted as $\mathbf{A}_i \in \mathbb{R}^K: \mathbf{A}_i \geq 0, \mathbf{A}_i^T\mathbb{1} = 1$, is a probability distribution that describes the probabilities of transitioning to one of the *K* possible hidden states, given that the chain is in state *i*, and each element $\mathbf{A}_{ij}$ represents the probability of transitioning from state *i* to state *j*. The observations, that depend on the hidden states, are specified by the emission distributions $p(\mathbf{x}_n|\mathbf{z}_n, \phi)$, where $\phi$ is the set of parameters that rule the distribution, that can be either discrete or continuous. Thus, a HMM is then completely specified by the set of components $\theta = (\mathbf{A}, \boldsymbol{\pi}, \phi)$ that must be learnt during the training phase.

As previously described, in this work, a Bayesian approach is adopted, in which current statistical procedures depend on the likelihood distribution of the models. Hence, the likelihood distribution $p(\mathbf{X}|\theta)$ of HMMs is needed.

*2.2.1 Likelihood distribution*

The likelihood distribution $p(\mathbf{X}|\theta)$ describes the joint probability of the sequence of observations $\mathbf{X} = \{\mathbf{x}_1, \mathbf{x}_2, ..., \mathbf{x}_N\}$ conditioned on the set of parameters $\theta$ and it is given as follows [17]:

$$p(\mathbf{X}|\theta) = \sum_{\mathbf{z}_N} \alpha(\mathbf{z}_N), \quad (3)$$

where

$$\alpha(\mathbf{z}_n) = p(\mathbf{x}_n|\mathbf{z}_n, \phi) \sum_{\mathbf{z}_{n-1}} \alpha(\mathbf{z}_{n-1}) p(\mathbf{z}_n|\mathbf{z}_{n-1}, \mathbf{A})$$

and

$$\alpha(\mathbf{z}_1) = p(\mathbf{z}_1|\boldsymbol{\pi}) p(\mathbf{x}_1|\mathbf{z}_1, \phi).$$

*2.2.2 Predictive distribution*

In this work, another quantity of interest is the predictive distribution $p(\mathbf{x}_{N+1}|\mathbf{X}, \theta)$, in which the observed data $\mathbf{X} = \{\mathbf{x}_1, \mathbf{x}_2, ..., \mathbf{x}_N\}$ is given and the goal is to predict the next observation $\mathbf{x}_{N+1}$. This distribution is given by [17]:

$$p(\mathbf{x}_{N+1}|\mathbf{X}, \theta) = \frac{1}{p(\mathbf{X}|\theta)} \sum_{\mathbf{z}_{N+1}} p(\mathbf{x}_{N+1}|\mathbf{z}_{N+1}, \theta) \cdot \sum_{\mathbf{z}_N} p(\mathbf{z}_{N+1}|\mathbf{z}_N, \theta) \alpha(\mathbf{z}_N). \quad (4)$$

However, during a conjunction event, 3 CDMs are received, on average, per day [7], and since the defined cut-off time is 2 days before the TCA, it could be advantageous to predict the information contained in the next $k$ collision warnings after the last released CDM. Future work may test the performance of predicting the next $k$ observations of an event but, in this work, this step is simplified, and only $\mathbf{x}_{N+1}$ is predicted and is used to benchmark the performance of HMMs.

## 3. Data

### 3.1 Data Analysis

As previously mentioned, the dataset used in this work is the one released by ESA during the CAC, which consists of CDMs collected by the ESA Space Debris Office between 2015 and 2019, in support of collision avoidance operations in the LEO region. Each row of the dataset corresponds to a single CDM, containing a total of 162 634 samples/CDMs (with 103 parameters each) and 13 154 unique conjunction events. The CDMs are identified by an event ID and data messages from the same conjunction event are grouped under the same identifier. Hence, each event represents a time series of CDMs that typically covers one week leading up to the TCA. Note that the values of all parameters contained in each CDM are propagated to the TCA.

However, not all events contained in the dataset are eligible for the ML approach, since spacecraft operators need time to make a decision regarding the performance of an avoidance manoeuvre. Thus, the events must follow some constraints [7]:
  (i) the events must have at least 2 CDMs, one to learn and one to use as label;
  (ii) the first CDM has to be released before the cut-off time (2 days until the TCA);
  (iii) the last CDM has to be released within 1 day of the TCA.

In total, the dataset contains 4904 events (approximately 37.2%) that do not satisfy the CDM requirements.

Since the goal of ML models in the collision probability assessment is to analyse the sequence of the values of the collision risk (the base-10 logarithm of the collision probability) contained in the CDMs received until the cut-off time and correctly identify whether an event is of high or low risk of collision, the data can be divided into two categorical classes, based on the risk that is present in the last CDM released in each event: if the risk is lower than $-6$, the event is considered of low risk, otherwise it is considered a high-risk event. This was the threshold defined during the CAC [7] and, since the same dataset is used and in order to better compare the obtained results, the same threshold is chosen in this work. Thus, in the dataset there are 12789 low-risk events (97.23%) and only 365 high-risk conjunctions (2.77%), highlighting the rare occurrence of high-risk events. The data imbalance problem poses to be the main challenge in collision risk estimation using ML methods. It is also important to note that the values of the risk contained in the CDMs are truncated at a lower bound of $-30$, in other words, the probability is truncated at $10^{-30}$. A closer look into the data showed that 63.5% of the total number of conjunction events has a final risk of $-30$, i.e., represent false alerts.

By performing an exploratory data analysis on the datasets, some anomalies can be found. There are parameters that contain extreme outliers or even physically impossible values – for example, negative ballistic coefficients or energy dissipation rates. In addition, in some collision warnings, the position standard deviations (along-track, radial, and transverse) of the target and chaser take values larger than the Earth radius, which is unrealistic from a physical point of view and affects the value of the collision probability.

### 3.2 Data Cleaning

During the data cleaning phase, in this work, the dataset is kept as close to the original as possible, in order to benchmark the performance of HMMs with data that is representative of real collision avoidance missions. In this phase, the CDMs that contain unrealistic or physically impossible values (like negative ballistic coefficients) are removed. Additionally, some parameters contain extreme outliers, like the position standard deviations of both objects, that have a maximum value greater than ten times the radius of the Earth. Accianiri et al. [18] also identified this problem and defined reasonable upper thresholds for the position standard deviations. In this work, the same upper thresholds are considered and the CDMs containing position errors above those values are removed. Furthermore, the events that do not follow the previously described constraints are also discarded.

The data cleaning described in this section results in a total of 5 917 events and 44 399 CDMs being removed from the dataset, which ends up with 7 187 (99.3%) low-risk events and only 50 (0.7%) high-risk conjunctions. Dealing with such imbalanced data poses to be the major challenge of this work.

### 3.3 Data Preparation and Setup

In collision avoidance, it is extremely important to identify high-risk conjunctions in order to prevent catastrophic collisions between space objects that can damage and lead to the destruction of operational spacecraft. However, as previously mentioned, the dataset is extremely imbalanced with only a very small percentage of the events having a final risk higher than $-6$. To mitigate this problem a strategy is followed. It was verified that when the naive forecast predicts the final risk value of an event as $-30$ (the risk values are truncated at that lower bound), 99.34% of the predictions are correct without any misclassification of high-risk events (after the removal of CDMs containing extreme outliers and parameters with impossible values). Thus, in this work, it is assumed that the $-30$ predictions by the naive forecast are trustworthy and, consequently, it is considered that those events do not require the application of ML models. So, those events are not used for training and, during the test phase, are directly predicted with a final risk of $-30$. With this approach, a significant amount of low-risk conjunctions are removed, which can help deal with the data imbalance problem, and the volume of training data is reduced, resulting in lower memory requirements and a lower computational time.

Furthermore, since the goal of this work is to benchmark the performance of Bayesian HMMs in risk estimation, the HMM will only learn the evolution of one single feature of the dataset – the risk. In other words, only the risk sequences contained in the CDMs will be analysed and predicted by the HMM. This way, this work provides a foundation for future research regarding the implementation of HMMs, with Bayesian statistics, in collision risk estimation.

Additionally, it is necessary to setup the data in order to be analysed by the model. The CDMs of each event are arranged in descending order regarding the time to the predicted TCA, and only the risk parameter is considered. Then, a stratified split is performed in order to preserve the same proportion of samples of each class, dividing the data into train and test sets using a ratio of $80:20$. The test set is only used at the end to evaluate the performance of the final model and, in each test event, only the CDMs released before the cut-off time can be used as input of the models, in order to simulate real-life operations, in which ML algorithms must predict the risk of collision with the available information until 2 days of the TCA. The training data is used to infer the parameters of the HMM.

However, at this point, a challenge arises. To infer the parameters of each model, current MCMC samplers require the evaluation of the log-likelihood density at each set of observations for each proposed set of parameters $\theta$ to be sampled. But, each event has a different number of CDMs, hence, to obtain the log probability of the model, it would be necessary to separately compute, in a loop, the logarithm of equation (4) for every set of observations of each event and then sum the result to obtain the joint log probability of the model. This would make the training of the model extremely slow and inefficient since this process would have to be repeated for every $\theta$ to be sampled. A solution is to vectorize the sequences of CDMs and compute the log-likelihood density for each sequence at once and then sum the result. To vectorize the sequences of collision warnings, an approximation must be done regarding the data setup of the training set. Typically, in real collision events, 3 CDMs are released per day [7] during the week leading up to the TCA, where the latest CDM available is always considered the best knowledge about the outcome of the close approach. Thus, an approach to ensure that all input sequences (events) have the same number of observations (CDMs) is to verify whether 3 CDMs are received each day and, if less than 3 collision warnings are received, the latest CDM received is repeated until there are 3 on that day. If there are no CDMs received prior to that day, the first observation received is repeated. This process is done for all days during the week leading up to the TCA and, after this, the events that don't match the highest number of observations are, again, manipulated by repeating the first released CDM.

The data setup process is schematized in Figure *4*.

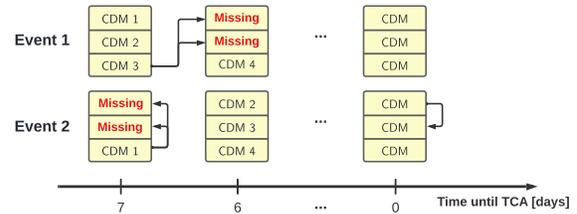

Figure 4. Data Setup schematization.

In summary, a schematic representation of the learning and prediction process is presented in Figure *5*, where $\hat{r}_{baseline}$, $\hat{r}_{HMM}$ and $\hat{r}$ denote the baseline, the HMM and final risk predictions, respectively.

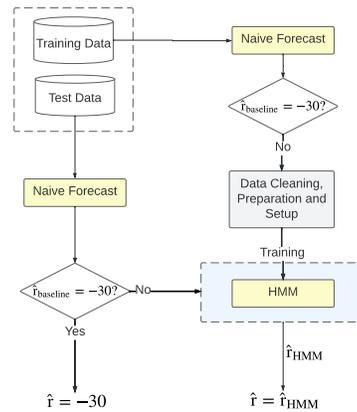

Figure 5. Learning and prediction procedure.

## 4. Bayesian Models

After the data cleaning and preparation, the training sequences of observations are used to infer the parameters of the models. In this work, it is believed that the risk/position errors generated by each latent variable of the HMM should be near a specific value and the occurrence of risk/position errors far from that value is less frequent. However, it is important to note that the variables to model must follow some constraints: the risk is truncated at a lower bound of $-30$ and cannot be greater than 0, because the risk is defined as the $log10$ of the collision probability; and the positional standard deviations are restricted to be greater than zero because they define the diagonal entries of the covariance matrix, which has to be positive semi-definite. To take these constraints into account, univariate Truncated Normal distributions are used as the emission distributions of the HMM, with lower and upper bounds of $-30$ and 0, respectively. Therefore, the parameters that must be inferred for the implemented model are the following, where K represents the number of possible hidden states:

(i) the transition probabilities represented by the matrix $\mathbf{A} \in \mathbb{R}^{K \times K}$: $\mathbf{A} \geq 0, \mathbf{A}\mathbb{1} = \mathbb{1}$;
(ii) the initial probability distribution represented by the vector $\boldsymbol{\pi} \in \mathbb{R}^K$: $\boldsymbol{\pi} \geq 0, \boldsymbol{\pi}^T\mathbb{1} = 1$;
(iii) the mean values $\boldsymbol{\mu} \in \mathbb{R}^K$ of the emission distributions;
(iv) the standard deviations $\boldsymbol{\sigma} \in \mathbb{R}_+^K$ of the emission distributions.

The parameters $\boldsymbol{\mu}$ and $\boldsymbol{\sigma}$ denote the set of mean values and standard deviations of the emissions, respectively, as $\boldsymbol{\mu} = [\mu_1, \mu_2, ..., \mu_K]$ and $\boldsymbol{\sigma} = [\sigma_1, \sigma_2, ..., \sigma_K]$, in which $\mu_k$ and $\sigma_k$ represent the mean and standard deviations of the Truncated Normal emission generated by the hidden state $k$. To find the best value for K, in each approach, a stratified cross-validation with 5 folds is performed.

As discussed in Section 2.1, in order to define the probabilistic models using a Bayesian approach, the following distributions are needed: the likelihood and the priors. The likelihood for HMMs is already defined in equation (3), leaving the definition of the priors. Thus, to perform Bayesian inference on the HMMs, it is essential to define the prior distributions for each of the parameters $\theta = (\mathbf{A}, \boldsymbol{\pi}, \boldsymbol{\mu}, \boldsymbol{\sigma})$ of the model.

Notice that the parameters of the priors presented in this section were chosen via a trial-and-error process (taking into consideration the constraints of the variables), so they are not unique and can be further improved.

*4.1 Priors for the HMM*
Regarding the parameters $\boldsymbol{\pi}$ and $\mathbf{A}_i \in \mathbb{R}^K$ : $\mathbf{A}_i \geq 0$, $\mathbf{A}_i^T\mathbb{1} = 1$ that define the initial state distribution and the rows of the transition matrix (with i ∈ {1, 2, ..., K}) a natural choice of priors is the Dirichlet distribution, that is confined to a simplex, i.e., all elements of the random variable belong to the interval [0,1] and sum up to one. The Dirichlet distribution is parameterized by the vector $\alpha \in \mathbb{R}^K$: $\alpha > 0$ whose elements must be positive real numbers and, in the case where all elements of α are equal to one, the distribution is equivalent to a uniform distribution over the simplex. In this work, there is no prior knowledge about the first state that generates the risk in each event nor about the hidden state transitions, so a Dirichlet distribution with elements of α equal to one is used as prior for $\boldsymbol{\pi}$ and $\mathbf{A}_i$.

Since the risk can only take values between $-30$ and 0, the mean values of the emission distributions are also restricted to be within the $-30$ to 0 range, so Truncated Normal distributions are also used as the prior distributions of $\boldsymbol{\mu}$, with lower and upper bounds of $-30$ and 0, respectively. To have good coverage of all the possible values that the observations can take, the mean of the prior distributions for the elements of μ are equally spaced within the range of $-30$ to 0 and the standard deviations are set to 4. For example, if $K = 3$, the priors for the elements of $\mu$ will be: $\mu_1 \sim \mathcal{TN}(\mu = -30, \sigma = 4)$, $\mu_2 \sim \mathcal{TN}(\mu = -15, \sigma = 4)$ and $\mu_3 \sim \mathcal{TN}(\mu = 0, \sigma = 4)$, in which the values of the lower and upper bounds of the Truncated Normal distribution are not shown, because these are fixed throughout.

As for the priors of $\boldsymbol{\sigma}$, it is necessary to choose a distribution that can only take positive values, because standard deviations are constrained to be greater than zero. The chosen distribution for the priors of the elements of $\boldsymbol{\sigma}$ is the inverse gamma distribution with parameters $\alpha$ and $\beta$ equal to 40 and 80, respectively (these values were chosen through a trial and error process).

In summary, the priors for the HMM are given by:

$$\pi \sim \mathcal{D}ir(\alpha = \mathbb{1}); \quad (5)$$
$$\mathbf{A}_i \sim \mathcal{D}ir(\alpha = \mathbb{1}), \quad \forall\, i \in \{1, ..., K\}; \quad (6)$$
$$\boldsymbol{\mu} \sim \mathcal{TN}(\mu = m, \sigma = 4, L = -30, U = 0); \quad (7)$$
$$\sigma_i \sim \mathcal{IG}(\alpha = 40, \beta = 80), \forall\, i \in \{1, ..., K\}; \quad (8)$$

in which $m \in \mathcal{M}_K(-30, 0)$, where $\mathcal{M}_K(a, b)$ is the set of $K$ evenly spaced numbers between $a$ and $b$. In addition, $L$ and $U$ denote the lower and upper bounds of the distributions, respectively, and $\mathbb{1}$ denotes a $K$-dimensional vector with all the elements equal to one.

*4.2 Inferences*
With the likelihood and prior distributions, it is possible to infer the parameters of the implemented models, using the NUTS. As previously described, a stratified cross-validation with five folds is performed in order to find the best value for K for the HMM and the best model then trained using the entire training set. During cross-validation, 3 chains of 2000 iterations are sampled for each model and, for the inference of the final HMM on the entire training set, 5 chains of 2000 iterations are sampled. The number of warm-up/tuning iterations per chain is set to 1000 and, after sampling, the samples used for tuning in each chain are discarded. In each sampling procedure, the target acceptance rate is set to a value of 0.8.

In this work, after sampling, it is necessary to deal with the label switching problem [19] — the label of the parameters switch between or within chains, due to the invariance of the likelihood and priors in the permutations of $\theta$. In this work, this is solved by relabelling the chains according to statistical analysis. Then, the convergence and autocorrelation of the sampled chains of each model are checked by visualizing the trace plots and by analyzing some of the convergence diagnostics criteria provided by PyMC, such as the Potential Scale Reduction ($\hat{R}$) [20] and the Effective Sample Size (ESS) [21]. If the inferences pass all the requirements, the samples of the posterior distribution can be used to obtain predictions.

## 5. Results and Discussion

This section presents the results obtained with the proposed model, comparing the predictions with the naive solution.

With the inferred posterior distribution, predictions for the HMM for the risk evolution can be obtained, using the predictive distribution shown in (4). In order to propagate the posterior uncertainty into the predictions, 400 random draws for the parameters of the model are taken from the inferred posterior distribution and are given as input to the predictive distribution (in this work, the performance of HMMs is benchmarked by predicting only $\mathbf{x}_{N+1}$), outputting a distribution that reflects the prediction uncertainty. The final predicted value of the risk of each event is the mean of the corresponding distribution. For each drawn set of parameters, a different predictive distribution is obtained and a random draw is taken from each of them. However, since truncated normal distributions are used as the emission distributions of the HMMs, the computation of the predictive distribution would require the sum of the probability distribution of multiple truncated normal distributions, resulting in a multimodal probability density function that is very difficult to compute analytically. Thus, this distribution is approximated by a truncated normal distribution with mean value equal to the first momentum of the distribution of (4) and variance equal to the second momentum of the distribution. Future work should focus on computing the true predictive distribution, without any approximation, but, in this work, this is simplified.

The metrics used to evaluate the models are the root mean squared error (RMSE), the mean absolute error (MAE), precision, recall, and F1 and F2 scores. In addition, confusion matrices are also used.

### 5.1 Model Results

To choose the best number of possible hidden states $K$, cross-validation is performed and $K$ is iterated between 4 and 10 states. For a lower number of $K$, it is considered that the HMMs have poor coverage of all the possible values of the desired parameter, and, for a higher number of $K$, the chains start converging into different values, indicating that the posterior distribution is multimodal. In the cases of multimodal posterior distributions, some of the parameters of the HMC/NUTS algorithms (like the mass matrix and the leapfrog step-size [16]) may only be locally optimized for one sharp density curvature of the posterior (one of the modes), during the warm-up/tuning phase, and the NUTS sampler can get stuck in that sharp region of the posterior density while sampling, failing to explore the rest of the density areas. Thus, by sampling randomly initialized chains, the sampler may get stuck in different modes, each time, which justifies the fact that the chains converge into different values. Future work may tackle this issue by using/developing an efficient sampler that can handle multimodality, but, in this work, this step is simplified. Note that only 3 chains are sampled during cross-validation, due to the large computing time during Bayesian inference, so it is possible that, even if the chains converge, the sampler may only be exploring part of the posterior distribution. Although this is not ideal, it still offers good information regarding the posterior distribution, since it explores the density regions near a mode of the desired distribution, in contrast to the maximum likelihood estimation or maximum a posteriori that only provide point estimates.

After cross-validation, the performance of the best model (in this case, the HMM with 8 states is chosen) is then tested using the test set (recall that the events with a naive forecast of $-30$ are directly predicted as having a final risk of $-30$). Table *1* shows the performance metrics for both the complete model and the baseline predictions.

Table 1 Performance metrics and confusion matrix for the implemented model and baseline solution.

| | **Metrics** | | | | | |
|---|---|---|---|---|---|---|
| Model | RMSE | MAE | Prec. | Recall | $F_1$ | $F_2$ |
| HMM | 0.82 | 1 | 0.11 | | 317.8 | |
| Naive | 0.007 | 0.01 | 0.093 | | 0.048 | |
| **Confusion Matrix** (Model \| Baseline) | | | | | | |
| | Pred. Low-Risk | | | Pred. High-Risk | | |
| True Low-Risk | 1406 \| 1398 | | | 33 \| 41 | | |
| True High-Risk | 3 \| 3 | | | 7 \| 7 | | |

The implemented model outperforms the baseline solution in all metrics (except for the recall). The results show that, despite the approximations made to build the model, the complex behaviour of the risk updates within the events, the data imbalance problem, and the fact that only one feature is used, the implemented model manages to outperform the naive forecast, which is considered a very strong predictor for this problem.

Table *1* also shows that both models have the same number of false negatives (true high-risk events miss-classified as low-risk) and true positives (true high-risk events correctly classified as high-risk), but the implemented model reduces the number of false positives by approximately 19.5%, which justifies its higher precision. The miss-classified low-risk events are the same for both models, but, to analyse the source of these errors, more data would be needed, since that three events are not a sufficiently large sample to take conclusions from. The events that were wrongly classified as high-risk by the implemented model were also miss-classified by the baseline and the evolution of some of these events (that are a good representation of the evolution of the events miss-classified as the positive class) is shown in Figure *6*.

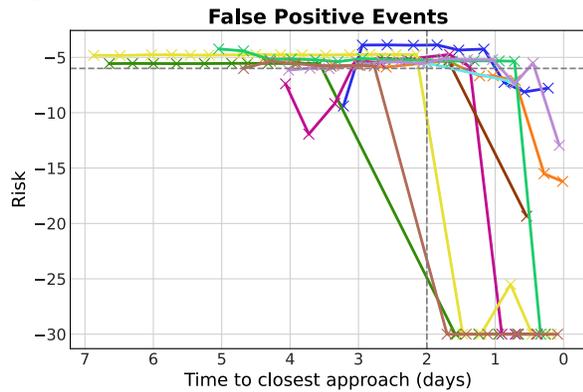

Figure 6. Risk evolution of the events that were wrongly classified as high-risk. The coloured lines represent different events and the crosses mark the risk updates.

All the false positive predictions have origin in events that have a risk value higher than the high-risk threshold before the cut-off time and, then, experience a significant jump toward lower values. This type of evolution highlights the complex and unpredictable behaviour of the risk updates within the events and suggests that, to correctly predict these risk transitions, more features of the dataset should be analysed by the ML models.

Figure *7* shows the predicted values of the risk (in the y axis) against the true risk values (in the x axis).

The predictions tend to be arranged in 8 steps, which correspond to the sampled mean of the emission distributions of the HMM. It can also be seen a large over-prediction of the $-30$ events. All these over-predicted events share the same behaviour: the risk updates evolve at high-risk values, but, after the cut-off time, there is a big risk transition, from high-risk values to $-30$. Figure *8* shows the actual time series evolution of some of the events with a true label of $-30$ that were over-predicted by the implemented model and the baseline and that represent the typical evolution of the risk of the over-predicted conjunction events.

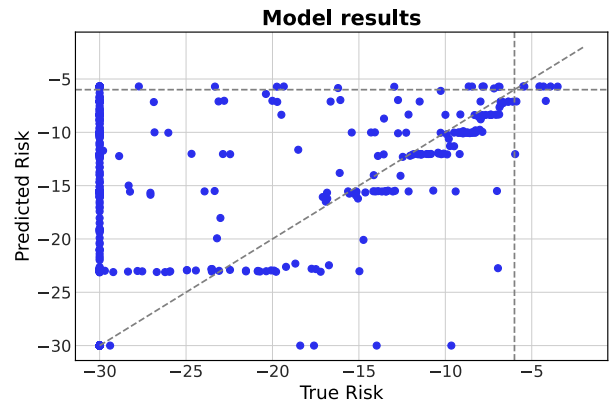

Figure 7. Predicted vs True risk values.

Figure *8* shows that most over-predicted $-30$ events evolve at high-risk values, but, after the cut-off time, they experience a big risk transition, that cannot be predicted, from higher risk values to $-30$. This type of risk evolution within the events also explains the high number of false positive predictions made by both models.

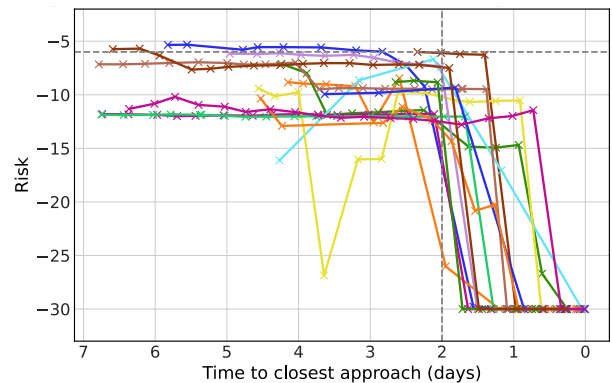

Figure 8. Typical time series evolution of the risk for over-predicted events with a true label of -30.

An examination of the predictions made by the implemented model and the baseline showed that the latter makes 218 big over-predictions of low true risk events, whereas the HMM only makes 172 over-predictions, where it is assumed that a large over-prediction of a $-30$ true risk event occurs when the corresponding predicted value is higher than $-30$. The implemented model reduces the number of over-predictions by approximately 21.1%, which shows that it has better performance than the baseline in identifying the risk transitions from higher to lower values.

As previously mentioned, an entire distribution is obtained for each prediction, so prediction intervals can be provided for each event. Figure *9* shows the 95%

Highest Density Interval (HDI) associated with each prediction for the true high-risk events of the test set. The prediction intervals have poor coverage of the true high-risk values and that the predictions seem to be "truncated" at an upper bound. It is important to highlight that this is a simple model that only analyses one feature of the dataset — the risk, which is extremely imbalanced. To improve the results, it could be beneficial to train the model with a larger dataset containing more high-risk conjunctions and explore the impact of other features.

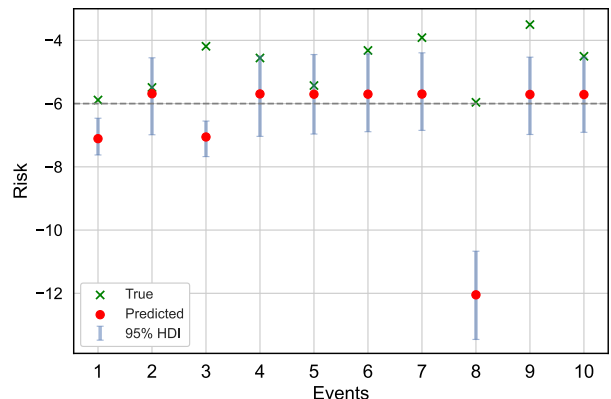

Figure 9. Representation of the true risk values of each event (green), the HMM predictions (red) and the 95% HDI area (area shaded in blue), for all the high-risk events contained in the dataset.

## 6. Conclusions

One of the conclusions taken from the CAC was that the naive forecast is a very strong predictor for the collision risk, indicating that the CDMs may follow the Markov property. This work tested this theory by benchmarking the performance of Bayesian HMMs by directly modelling and predicting the evolution of the risk of collision values contained in the CDMs of close approach events. The results have shown that the implemented model managed to outperform the baseline solution in all metrics, despite all the approximations made, the data imbalance problem, the fact that only the risk feature was used, and the complex behaviour of the risk updates within the events. These promising results further add to the idea that the CDMs may follow the Markov property and suggest that this method should be further explored. In addition, this work provides a foundation for future research regarding the implementation of Bayesian HMMs to the challenge of applying ML in collision avoidance.